\documentclass{article}

     \PassOptionsToPackage{numbers, compress}{natbib}

\usepackage[final]{neurips_2019_ml4ad}

\usepackage[utf8]{inputenc} 
\usepackage[T1]{fontenc}    
\usepackage{hyperref}       
\usepackage{url}            
\usepackage{booktabs}       
\usepackage{amsfonts}       
\usepackage{nicefrac}       
\usepackage{microtype}      
\usepackage{amsmath}
\usepackage{amssymb}
\usepackage{float}
\usepackage{gensymb}
\usepackage{dirtytalk}
\usepackage[numbers]{natbib}
\usepackage{graphicx}

\title{SoildNet: Soiling Degradation Detection in Autonomous Driving}

\author{%
  Arindam Das\\
  Detection Vision Systems\\
  Valeo India\\
  \texttt{arindam.das@valeo.com} \\
}

\begin{document}

\maketitle

\begin{abstract}
  In the field of autonomous driving, camera sensors are extremely prone to soiling because they are located outside of the car and interact with environmental sources of soiling such as rain drops, snow, dust, sand, mud and so on. This can lead to either partial or complete vision degradation. Hence detecting such decay in vision is very important for safety and overall to preserve the functionality of the \say{autonomous} components in autonomous driving. The contribution of this work involves: 1) Designing a Deep Convolutional Neural Network (DCNN) based baseline network, 2) Exploiting several network remodelling techniques such as employing static and dynamic group convolution, channel reordering to compress the baseline architecture and make it suitable for low power embedded systems with $\sim$1 TOPS, 3) Comparing various result metrics of all interim networks dedicated for soiling degradation detection at tile level of size $64 \times 64$ on input resolution $1280 \times 768$. The compressed network, is called SoildNet (\textbf{S}and, sn\textbf{O}w, ra\textbf{I}n/d\textbf{I}rt, oi\textbf{L}, \textbf{D}ust/mu\textbf{D}) that uses only 9.72\% trainable parameters of the base network and reduces the model size by more than 7 times with no loss in accuracy.
\end{abstract}

\section{Introduction}

\begin{figure*}[!htb]
\centering
\minipage{0.17\textwidth}
    \includegraphics[width=\linewidth]{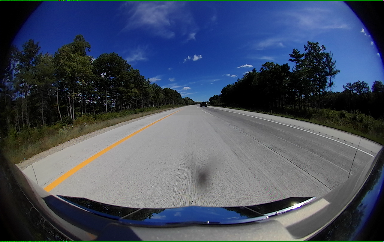}
    \hspace*{1cm}{$(a)$}\label{fig:FV}
\endminipage
\hspace{0.1cm}
\minipage{0.17\textwidth}
    \includegraphics[width=\linewidth]{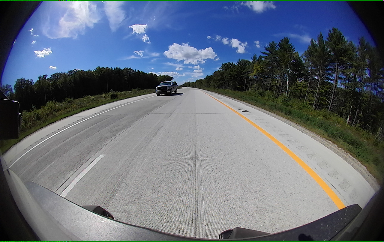}
    \hspace*{1cm}{$(b)$}\label{fig:RV}
\endminipage
\hspace{0.1cm}
\minipage{0.17\textwidth}
    \includegraphics[width=\linewidth]{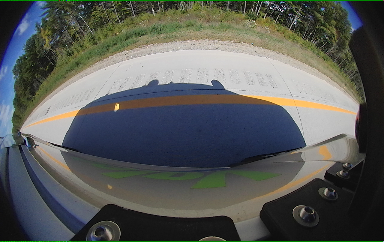}
    \hspace*{1cm}{$(c)$}\label{fig:MVL}
\endminipage
\hspace{0.1cm}
\minipage{0.17\textwidth}
    \includegraphics[width=\linewidth]{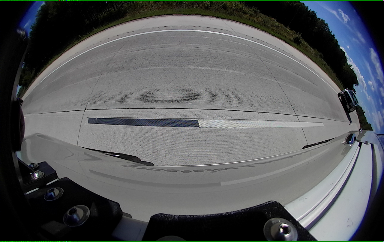}
    \hspace*{1cm}{$(d)$}\label{fig:MVR}
\endminipage
\caption{Different field of vision of surround-view cameras: $(a)$ Front, $(b)$ Rear, $(c)$ Left and \newline $(d)$ Right}\label{fig:camera_vision}
\end{figure*}

Vision based algorithms are particularly depending on the image data that are passed from the camera sensors with almost $360$\degree surrounding view as shown in figure \ref{fig:camera_vision}. The quality of the input vision needs a certain level of validation before being fed to other downstream algorithms since the performance of the allied processes degrade severely if there is substantial decay in the vision. Hence it is extremely critical to detect about the degradation in the input vision and report to the system while aiming for Level $4$ autonomous driving. This will ensure the safety of the passengers and others to avoid any unprecedented event.


There are very few works available in the literature on the reported problem statement and the approaches can be classified in two categories, $1)$ Image restoration and $2)$ Soiling detection. In the first category, there has been attempts to recover the input image by removing rain drops \cite{pfeuffer2019robust}. Another successful effort has been to dehaze \cite{ki2018fully} the high resolution ultrasound images. For both the approaches, the network was trained with a pair of \textit{defective-clean} images. However, the deployability of the techniques in real-time on a low power automotive SoC is questionable. In the second category, in \cite{uricar2019yes} GAN (Generative Adverserial Network) was used to augment the soiling samples. The same approach was followed in \cite{uricar2019soilingnet} as well.

\begin{figure*}[!htb]
\centering
\minipage{0.17\textwidth}
    \includegraphics[width=\linewidth]{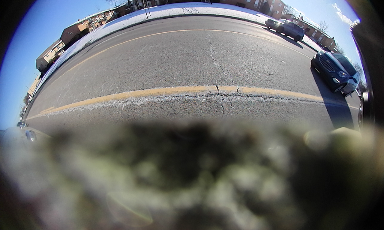}
    \hspace*{1cm}{$(a)$}\label{fig:grass}
\endminipage
\minipage{0.17\textwidth}
    \includegraphics[width=\linewidth]{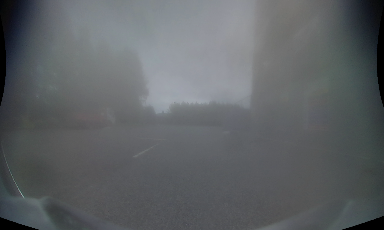}
    \hspace*{1cm}{$(b)$}\label{fig:fog}
\endminipage
\minipage{0.17\textwidth}
    \includegraphics[width=\linewidth]{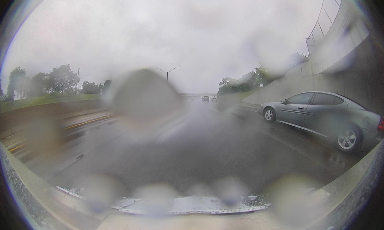}
    \hspace*{1cm}{$(c)$}\label{fig:rain_drops}
\endminipage
\minipage{0.17\textwidth}
    \includegraphics[width=\linewidth]{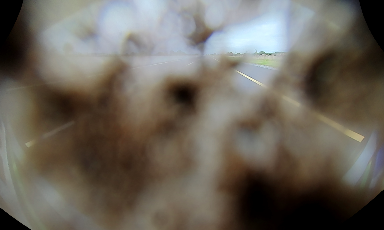}
    \hspace*{1cm}{$(d)$}\label{fig:dirt}
\endminipage
\minipage{0.17\textwidth}
    \includegraphics[width=\linewidth]{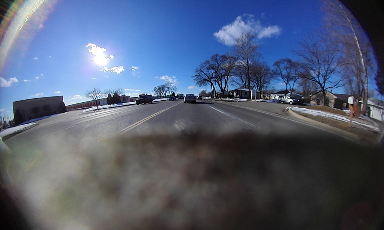}
    \hspace*{1cm}{$(e)$}\label{fig:mud_splash}
\endminipage
\minipage{0.17\textwidth}
    \includegraphics[width=\linewidth]{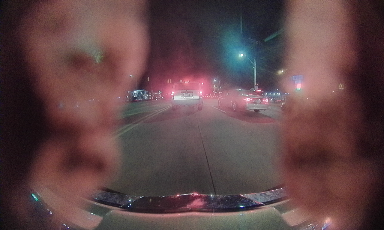}
    \hspace*{1cm}{$(f)$}\label{fig:mud_spalsh_night_mode}
\endminipage
\caption{Different types of \textit{soiling}:  $(a)$ grass, $(b)$ fog, $(c)$ rain drops, $(d)$ dirt, $(e)$ splashes of mud, $(f)$ splashes of mud in night}\label{fig:soiling_types}
\end{figure*}

\section{Contribution}

\begin{itemize}
\item[--] \textbf{Critical but less explored problem}: Soiled vision can impede other vision based applications and cause several safety issues if it goes unnoticed. Hence this important problem requires a fast and \textit{low resource environment} friendly solution. However, the availability of very few works in the literature demonstrates that this task is quite unexplored. To the best of author's knowledge, this is the first study on tile based soiling degradation detection where the network recommendations have been investigated meticulously from the embedded platform perspective.
\item[--] \textbf{Network optimization and analysis}: In this study, it has been demonstrated how the main take aways of few of the existing network remodeling methods \cite{xie2017aggregated} \cite{zhang2018shufflenet}  can help to optimize a base network incredibly by 7 folds in model size and use only 9.72\% trainable parameters of the base network with no sign of loss in accuracy. Eventually the best optimized network outperforms all the other incrementally optimized variants of the base network for soiling detection task. However, it is to be noted that in any of the earlier works, group convolution was never applied at all convolution layers due to the issue with insufficient feature blending. The present study overcomes this bottleneck and discusses in section 5.
\item[--] \textbf{Tile level soiling detection}: Most of the times, soiling effect appears discontinuously over the image. Hence it makes more sense to detect soiling more locally than at the image level. Detecting soiling more locally does not completely disqualify the input image to be used for other algorithms such as VSLAM (Visual Simultaneous Localization and Mapping) \cite{karlsson2005vslam}, motion stereo \cite{unger2014parking}, environment perception \cite{franke20056d}, $3$D object detection \cite{chen2016monocular}, semantic segmentation \cite{das2019design}. Rather all the downstream algorithms can still run on the tiles that are predicted as soiling free. It should be noted that the definition of tile in this paper is not limited to the defined dimension, rather it can be extended to as far as the whole image and reduced as much as pixel level.
\end{itemize}

\section{Soiling Degradation Detection}
With reference to the detail explanation in earlier sections, while it is clear that there is no way to protect the camera sensors from being effected by the various sources of degradation. It is important for obvious reasons to recover the visual field when degradation is detected on the substantial portion of the input image. It is the cleaning system that is invoked automatically to remove the soiled objects by spraying warm water. This system includes a separate tank that reserves water for this purpose and needs refuelling just like gas. System level details on the cleaning system is shown in \cite{uricar2019soilingnet}. Various types of the soiling and its classes that are considered in this experiment are discussed below.

\subsection{Types of Soiling}
The decline in vision can be either due to adverse weather conditions and this covers soiling types for example  \textit{snow}, \textit{rain drops}, \textit{fog} etc. or the other types that emerge regardless bad weather such as \textit{mud}, \textit{grass}, \textit{oil}, \textit{dust}, \textit{sand} etc. Figure \ref{fig:soiling_types} shows few examples of different soiling types that are considered in this experiment.

\subsection{Classes of Soiling}
Different types of soiling discussed in the previous section are divided into three categories: \textit{clean}, \textit{opaque} and \textit{transparent}. This categorization is done based on the visibility within the region of interest that is per tile.

\begin{itemize}
\item[--] \textbf{Clean}: When a tile has completely freeview then it is categorized as clean. 

\item[--] \textbf{Opaque}: A tile is marked as opaque when the vision is totally blocked. The complete decay in vision can happen because of any type of soiling that is discussed in the previous section. 

\item[--] \textbf{Transparent}: Sometimes, due to the uneven distribution of the soiling objects on the camera lens, majority of the tiles in the input image do not loose complete visibility. Rather one can see through few of the tiles that are partly affected. Tiles with such qualities are considered as \textit{transparent} in this study.
\end{itemize}

It is possible that presence of multiple classes are observed within one tile, however the tiles are annotated based on the presence of the dominant class per tile. In this study, an input image of resolution $1280 \times 768$ is annotated per tile of size $64 \times 64$ and each tile represents a soiling class, hence it is possible to see an input image that contains all soiling categories across tiles.

\section{Dataset}

Due to unavailability of any public dataset, several driving scenes or videos have been used to extract the frames, however not all successive frames were extracted. This is because highly correlated samples do not contribute much during training.

For the reported problem, it is more critical to predict a \textit{clean} tile correctly. This is because, a high number of false positives will lead to cleaning a camera that is already clean more frequently. As an effect the water tank will need refuelling repeatedly. Hence it makes sense to make the model moderately biased towards the \textit{clean} class. In this experiment, total $144\,053$ sample images are used out of which $70\,000$ samples are pure clean images, which means that all tiles are soiling free. Higher number of clean samples will help to learn better discriminative features of clean class, hence the model tends to be biased towards clean. The distribution of sample across cameras is as follows, FV: $36\,259$; RV: $36\,160$, MVR: $35\,435$; MVL: $36\,199$. A tile of size $64 \times 64$ on input resolution $1280 \times 768$ makes $20$ tiles along width and $12$ tiles along height, thus a single sample contains total $20 \times $12 tiles and the tile based class distributions are - Clean: $25\,459\,238$; Opaque: $6\,341\,435$; Transparent: $2\,772\,047$. Change in camera view impacts the appearance of the object significantly, and thus it is necessary to check how the classes are well spread at tile level across different camera views. Table \ref{class distribution} shows that the classes are adequately distributed in all four camera views. 

\begin{table}[h!]
\centering
\begin{tabular}{||c c c c c ||}
 \hline
 Class & FV & RV & MVR & MVL \\ 
 \hline\hline
 Clean & 36141 & 35926 & 35435 & 35755 \\ 
 Opaque & 17877 & 17813 & 16740 & 17764 \\
 Transparent & 17866 & 17716 & 17648 & 17753 \\ [1ex]
 \hline
\end{tabular}
\vspace{0.1cm}
\caption{Camera view-wise presence of all three classes}
\label{class distribution}
\end{table}

The dataset is subdivided into training, validation and test sets with partition ratios of $60$\%, $20$\% and $20$\% respectively. In Figure \ref{fig:soiling_types}, a few samples are shown from the dataset. It is to be noted that all the samples are fisheye and in YUV420 \cite{yuan2004color} planar image format as produced by the ISP (Image Signal Processor) cameras. For soiling detection task, fisheye images were not corrected because a separate preprocessing module would be required and that would increase the overall inference time to get the end-to-end results.

\begin{figure*}[!htb]
\centering
\minipage{1\textwidth}
    \includegraphics[width=\linewidth]{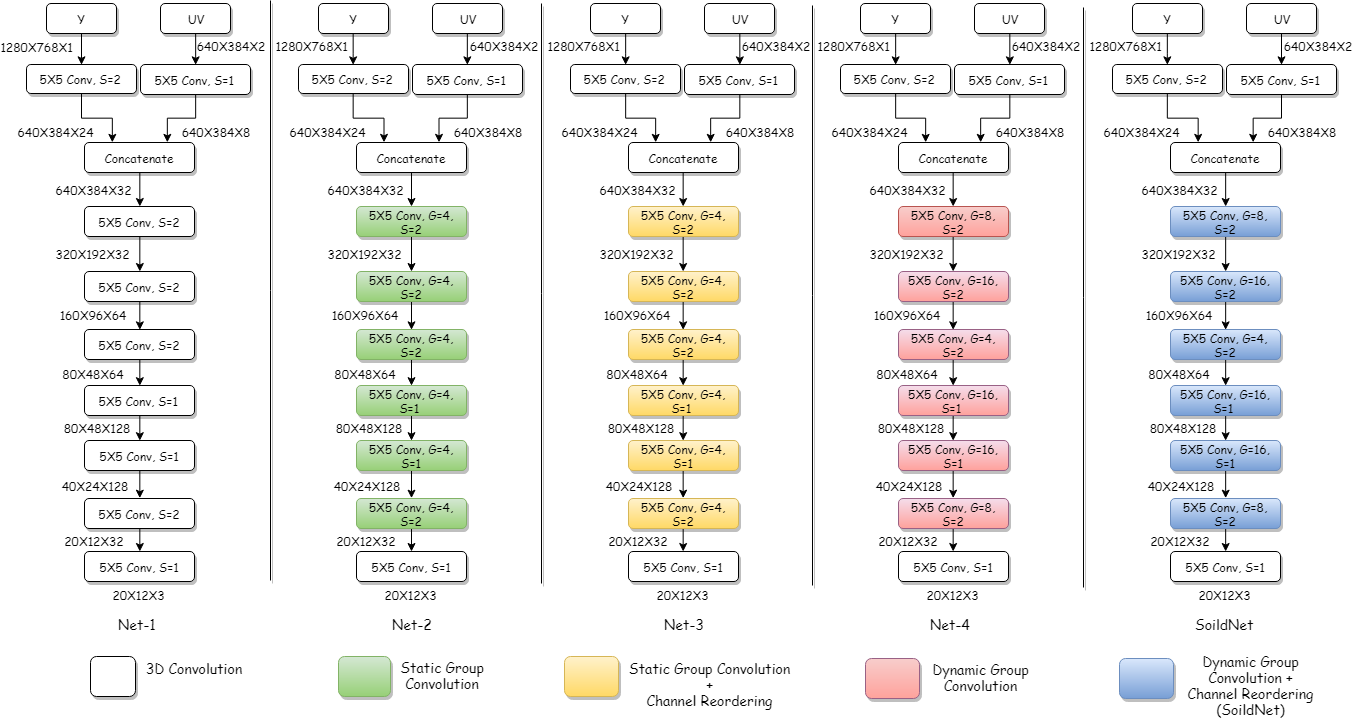}
    \caption{Proposed networks with different schemes for soiling detection. $Conv.:$ Convolution, $G:$ group size, $S:$ stride size}\label{all_nets}
\endminipage
\end{figure*}

\section{Proposed Method}
To accomplish soiling degradation detection task on low power automotive SoC, first a base network (Net-$1$) is designed to take input in YUV420 planar format where the dimension of Y, U and V channel are $1280 \times 768$, $640 \times 384$ and $640 \times 384$ respectively. As there is a mismatch of dimension between Y and UV, so the network is designed to take two inputs, one is for Y and the other one is UV together. Later through convolution operation, both set of feature maps (Y and UV) are brought down to similar dimension and concatenated to make the network single stream. This approach can be checked in details in \cite{boulay2019yuvmultinet}. Net-$1$ is further refactored to $4$ other networks (Net-$2$, Net-$3$, Net-$4$ and SoildNet) as shown in figure \ref{all_nets} to obtain the best variant of Net-$1$ that will be lightweight and more efficient. The network refactoring is done using group convolution and channel reordering that are discussed in the next section.

\subsection{Group Convolution}
The idea to perform convolution operation group wise was first introduced in AlexNet \cite{krizhevsky2012imagenet}. However the main intention was to distribute the number of operations in two GPUs. Later in ResNeXt \cite{xie2017aggregated}, this proposal was used to boost the accuracy with reduced network complexity on an object recognition task. The earlier work in \cite{xie2017aggregated} considered static number of groups in the network. The current work extends this concept by adding group convolution in all convolution layers (Net-$2$, Net-$3$, Net-$4$, SoildNet), also we experiment with dynamic group size (Net-$4$, SoildNet) to reduce the network complexity by more than two times in trainable parameters (Net-$3$ vs. Net-$4$). The network schemes do not contain residual connections because group convolution was found to be not very effective for the networks with low depth as presented in \cite{das2018evaluation}. Also, a similar study on residual connection for lightweight networks \cite{dasevaluation} is the reason not to use them in any of the proposals. While adding group convolution at all layers of the network brings another challenge of insufficient feature blending. This is overcome through channel reordering that is discussed in the following section.

\subsection{Channel Reordering}
The concept of channel reordering is highly inspired from ShuffleNet \cite{zhang2018shufflenet}. While performing group convolution, the feature information are limited within the group. To make the features blend across groups, the feature maps are shuffled in an ordered way that makes sure in the next layer each group contains at least a candidate feature map from each group of the previous layer.

In this experiment, two constraints have been found in ShuffleNet and they are solved in this study. First, it is now well known that group convolution is effective to bring down the network complexity but at the some time this method can not be applied in all convolution layers. This is because the feature information will not be spread across all feature maps and this step is necessary to learn better descriptors. The main reason of the insufficient feature blending is that all the feature maps will never undergo convolution operation together as they are separated by groups. After performing one or two layers of consecutive group convolution, generally a convolution layer with kernel size $1 \times 1$ is added to blend the features across channel. As an effect of this, convolution operation on all feature maps again shoots the number of trainable parameters significantly high. In order to execute group convolution at all convolution layers throughout the network, channel reordering is added that helps to ensure feature blending across groups. Certainly following this way, feature blending will not be as effective as normal convolution on all feature maps but definitely the blending will be mostly same as the network is trained for higher number of epochs. And the late convergence of the network with group convolution at all layers and channel reordering impact only on the training time. Another reason that the channel reordering was not applied at all layers in ShuffleNet due to its usage of residual connection.

ShuffleNet uses channel shuffling while maintaining the constant number of groups in the network. This significantly limits further reduction of the GMACS (discussed in the next section), number of parameters and model size considering group convolution is not performed at all layers. In this experiment, we designed two networks (Net-$4$ and SoildNet) such that Net-$4$ contains different number of groups at each layer and SoildNet contains same number of groups as Net-$4$ but it includes channel reordering method. Here, the group sizes are determined based on the following idea: One convolution layer with higher number of groups heavily reduces the number of trainable parameters but it limits feature blending due to the feature maps are separated by more number groups then next convolution layer should use less number of groups to blend the features well. So a good balance is maintained between reducing the number of trainable parameters and feature blending. In SoildNet, apart from group convolution with dynamic group size, channel reordering ensures that the features are blended even when the group is size less by shuffling the feature maps across groups. The effectiveness of this approach can be seen in table \ref{network_results_classwise}.

\newcommand{\specialcell}[2][c]{%
  \begin{tabular}[#1]{@{}c@{}}#2\end{tabular}}

\newcommand\Tstrut{\rule{0pt}{5.6ex}}       
\newcommand\Bstrut{\rule[-0.9ex]{0pt}{0pt}} 
\newcommand{\TBstrut}{\Tstrut\Bstrut} 

\section{Analysis of SoildNet}
CNN mostly follows two major operations while performing a convolution task - multiplication and addition. Total number of operations involved in a network is represented by GMACS (Giga Multiply Accumulate Operations per Second) unit. Table \ref{network_analysis} furnishes the details about the number of trainable parameters used in all five network schemes (Net-$1$, Net-$2$, Net-$3$, Net-$4$, SoildNet) along with their GMACS and model size. As an effect of more than $90$\% reduction of network parameters due to group convolution from baseline network in two variants of SoildNet (with and without channel reordering), the model size is reduced by more than $7$ times. This is quite a significant and encouraging number while deploying a model on a low power SoC. Also it is clearly seen that the GMACS of SoildNet (and Net-$4$) is quite less than the baseline or other network schemes. Thus this factor helps SoildNet to be faster during inference.

\begin{table}[!htb]
\centering
\renewcommand{\arraystretch}{1.5}
\scalebox{0.8}{
\begin{tabular}{||c c c c ||}
 \hline
 Network & \specialcell{Operations\\(GMACS)}  & \specialcell{Parameters} & \specialcell{Model size\\(KB)}  \\ 
 \hline\hline
 ResNet-10 & 24.19 & 4,937,881 & 68,261 \\
 Net-1 & 4.203 & 900,849 & 3,569  \\ 
 Net-2 & 1.236 & 228,401 & 965 \\
 Net-3 & 1.236 & 228,401 & 965 \\ 
 Net-4 & 0.6672 & 87,601 & 478 \\
 SoildNet & 0.6672 & 87,601 & 478 \\ [0.25ex]
 \hline
\end{tabular}}
\vspace{0.1cm}
\caption{Analysis of computation complexity of all network proposals including ResNet-$10$  \cite{he2016deep}}
\label{network_analysis}
\end{table}

It is to be noted that channel reordering technique does not have any influence on the size of network parameters because it only changes the position of the feature maps thus the number of parameters in the network remain same. For the very same reason, there is no impact on model size as well for the network with and without channel reordering.

As the present study is focused on designing efficient lightweight networks for soiling degradation detection task, it seemed interesting to do an analysis from the perspective of computation complexity of a standard network such as ResNet-$10$  \cite{he2016deep} (the lightest version of the ResNet family) for an embedded platform with $\sim$1 TOPS (Tera Operations per Second). Table \ref{network_analysis} shows the GMACS, number of trainable paramaters and model size of ResNet-10 with respect to the input resolution used in this work. However, due to the reasons stated below ResNet-$10$ is not considered further in this work.

\begin{itemize}
\item[--] \textbf{Number of operations}: The reported GMACS of ResNet-$10$ is way too much high to be considered for an embedded platform.
\item[--] \textbf{Model size}: Automotive SoCs generally provide only few megabytes of memory where all the autonomous algorithms of the ADAS system need to be accommodated. Hence, with such budget constraint, acceptance of a model of size more than $5$MB is questionable, especially when we target higher FPS (Frames Per Second).
\item[--] \textbf{Residual connections}: The memory budget heavily increases with the networks containing residual connections since the feature maps need to be saved in the memory to perform addition later at the end of the residual connection. During feature maps retrieval, DMA (Direct Memory Access) transfers the data from the storage to the very limited cache memory for feature map summation. If the cache memory fails to hold the entire data then DMA keeps on copying and processing the data partially. Eventually with more number of feature maps this rolling buffer method is performed quite a number of times and it leads towards higher inference time of the network on the embedded platform.

\end{itemize}

\section{Embedded Platform Constraints}
The network models explained in section $5$ follow floating point operations because these architectures are trained in Keras \cite{chollet2015keras} framework on GPU. However, most of the embedded SoCs follow 16-bit fixed point operations, hence the data are quantized when the model trained on a GPU is deployed on a target device. In this study, the throughput of the SoC is $\sim$1 TOPS and capable to support $400$ GMACS. 
All the network proposals are well aligned with the constraints of the CNN IP (Image Processor) to make sure its full utilization of the resources. For example, pooling layer is not used in the network to reduce an extra clock-cycle, rather stride is applied to reduce the problem space. Also all convolution kernels are $5 \times 5$ to ensure $100$\% core utilization of the CNN IP.

As per GPU implementation, performing group convolution involves first slicing input feature maps into a number of groups, then execute convolution operation on each group and finally concatenate the output feature maps of all groups. However this extra overhead does not exist in the embedded environment. This is because on CNN IP memory address of each feature map is passed while doing convolution operation, so to follow group wise convolution, simply memory address of the feature maps need to be sent in group wise fashion. It is also to be highlighted that the channel reordering needs extra effort on GPU that includes again feature map slicing in a way so that resultant feature maps are in desired order. However, this effort is completely neutralized on the hardware since the feature maps are handled only through memory locations. Apparently when the succeeding convolution layer would be expecting reordered feature maps then the feature maps from the desired memory locations would be sent to ensure channels are reordered. 

\section{Experimental Results}
This section provides details about the performance of all network propositions on the test dataset. The discussion includes training strategy that was followed for all $5$ networks and reporting classwise standard metrics for overall network evaluation.

\subsection{Training Strategy}

All the network schemes are implemented using Keras \cite{chollet2015keras} framework. Batch normalization layer is added between each convolution layer and ReLU as activation. Training was done batch wise with batch of size $16$ for $50$ epoch, initial learning rate was set to $0.001$ along with an optimizer Adam \cite{kingma2014adam}. Categorical cross entropy and categorical accuracy were used as loss and metrics respectively for all networks. As the networks are less in depth and no pre-training was done, to make the network weights more robust, the concept of layer-wise training in a supervised fashion could be adapted as presented in \cite{roy2016generalized}.


\subsection{Evaluation}
To execute a fair comparison about the efficacy of all network schemes, few standard metrics are considered such as TPR (True Positive Rate), TNR (True Negative Rate), FPR (False Positive Rate), FNR (False Negative Rate) and FDR (False Discovery Rate) respectively. In order to get better insight about the performance, these metrics are computed for each class on the test dataset. The rule to interpret these metrics is to aim for higher values of TPR, TNR and lower values of FPR, FNR, FDR respectively.

\begin{table*}[!htb]
    \begin{center}
    \large
    \scalebox{0.46}{
    \begin{tabular}{||c | c c c | c c c | c c c | c c c | c c c||}
    \hline
    - & 
    \multicolumn{3}{c|}{ True Positive Rate (TPR)} &
    \multicolumn{3}{c|}{ True Negative Rate (TNR)} &
    \multicolumn{3}{c|}{ False Positive Rate (FPR)} &
    \multicolumn{3}{c|}{ False Negative Rate (FNR)} &
    \multicolumn{3}{c||}{ False Discovery Rate (FDR)} \\  [2ex]
    \hline
    \hfil Network & Clean & Opaque & Transparent & Clean & Opaque & Transparent & Clean & Opaque & Transparent & Clean & Opaque & Transparent & Clean &  Opaque & Transparent \\ [2ex]
    \hline
    \hline
    \hfil \large Net-$1$ & \large 0.9607 & \large 0.9157 & \large 0.4939 & \large 0.8864 & \large 0.9602 & \large 0.9753 & \large 0.1135 & \large 0.0397 & \large 0.024 & \large 0.0392 & \large 0.0842 & \large 0.506 & \large 0.0402 & \large 0.1639 & \large 0.3632 \\ [1ex]
    \hfil \large Net-$2$ & \large \textbf{0.9902} & \large 0.8923 & \large 0.3706 & \large 0.8048 & \large \textbf{0.9835} & \large 0.9861 & \large 0.1951 & \large \textbf{0.0164} & \large 0.0138 & \large 0.0097 & \large 0.1076 & \large 0.6293 & \large 0.0652 & \large \textbf{0.0766} & \large 0.3001 \\ [2ex]
    \hfil \large Net-$3$ & \large 0.9724 & \large 0.921 & \large 0.5413 & \large 0.9024 & \large 0.9708 & \large 0.9759 & \large 0.0975 & \large 0.0291 & \large 0.024 & \large 0.0275 & \large 0.0789 & \large 0.4586 & \large 0.0343 & \large 0.1249 & \large 0.3371 \\ [2ex]
    \hfil \large Net-$4$ & \large 0.9916 & \large 0.9302 & \large 0.2859 & \large 0.8136 & \large 0.9739 & \large \textbf{0.9934} & \large 0.1863 & \large 0.026 & \large \textbf{0.0065} & \large \textbf{0.0083} & \large 0.0697 & \large 0.714 & \large 0.0624 & \large 0.1123 & \large \textbf{0.2087} \\ [2ex]
    \hfil \large SoildNet & \large 0.9556 & \large \textbf{0.9303} & \large \textbf{0.5973} & \large \textbf{0.938} & \large 0.9642 & \large 0.9649 & \large \textbf{0.0619} & \large 0.0357 & \large 0.035 & \large 0.0443 & \large \textbf{0.0696} & \large \textbf{0.4026} & \large \textbf{0.0224} & \large 0.1479 & \large 0.4019 \\ [1ex]
    \hline
    \end{tabular}}
    \label{table1}
    \vspace{0.1cm}
    \caption{Comparison of classwise accuracy between the base model (Net-1) and other network propositions (Net-$2$, Net-$3$, Net-$4$, SoildNet) for tile level soiling degradation detection}\label{network_results_classwise}
    \end{center}
\end{table*}

\begin{figure*}[!htb]
\centering
\minipage{0.13\textwidth}
    \includegraphics[width=\linewidth]{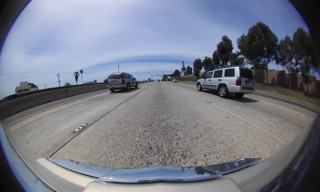}
\endminipage
\hspace{0.05mm}
\minipage{0.13\textwidth}
    \includegraphics[width=\linewidth]{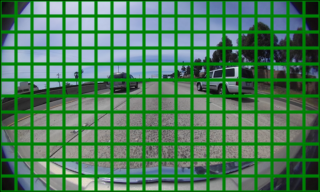}
\endminipage
\hspace{0.05mm}
\minipage{0.13\textwidth}
    \includegraphics[width=\linewidth]{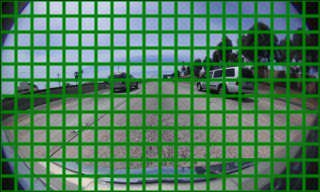}
\endminipage
\hspace{0.05mm}
\minipage{0.13\textwidth}
    \includegraphics[width=\linewidth]{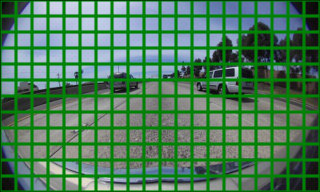}
\endminipage
\hspace{0.05mm}
\minipage{0.13\textwidth}
    \includegraphics[width=\linewidth]{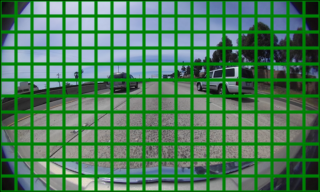}
\endminipage
\hspace{0.05mm}
\minipage{0.13\textwidth}
    \includegraphics[width=\linewidth]{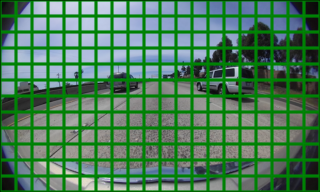}
\endminipage
\hspace{0.05mm}
\minipage{0.13\textwidth}
    \includegraphics[width=\linewidth]{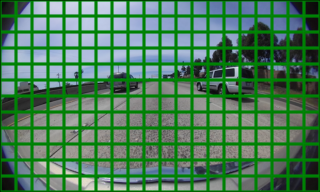}
\endminipage

\minipage{0.13\textwidth}
    \includegraphics[width=\linewidth]{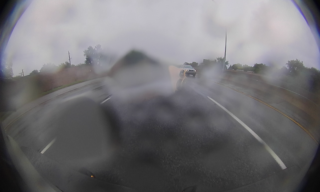}
\endminipage
\hspace{0.05mm}
\minipage{0.13\textwidth}
    \includegraphics[width=\linewidth]{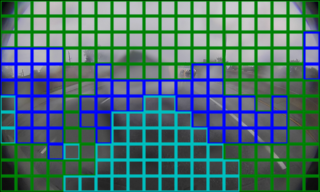}
\endminipage
\hspace{0.05mm}
\minipage{0.13\textwidth}
    \includegraphics[width=\linewidth]{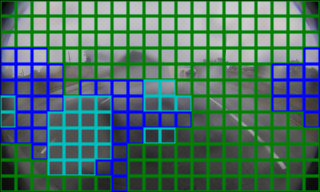}
\endminipage
\hspace{0.05mm}
\minipage{0.13\textwidth}
    \includegraphics[width=\linewidth]{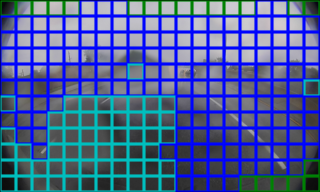}
\endminipage
\hspace{0.05mm}
\minipage{0.13\textwidth}
    \includegraphics[width=\linewidth]{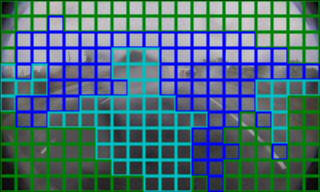}
\endminipage
\hspace{0.05mm}
\minipage{0.13\textwidth}
    \includegraphics[width=\linewidth]{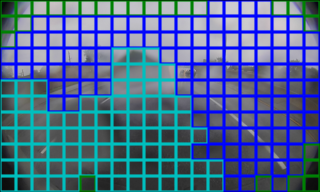}
\endminipage
\hspace{0.05mm}
\minipage{0.13\textwidth}
    \includegraphics[width=\linewidth]{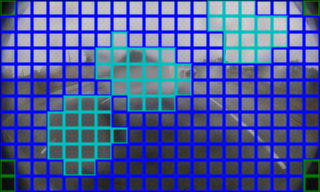}
\endminipage

\minipage{0.13\textwidth}
    \includegraphics[width=\linewidth]{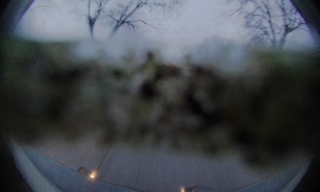}
\endminipage
\hspace{0.05mm}
\minipage{0.13\textwidth}
    \includegraphics[width=\linewidth]{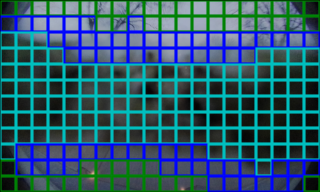}
\endminipage
\hspace{0.05mm}
\minipage{0.13\textwidth}
    \includegraphics[width=\linewidth]{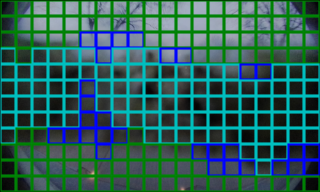}
\endminipage
\hspace{0.05mm}
\minipage{0.13\textwidth}
    \includegraphics[width=\linewidth]{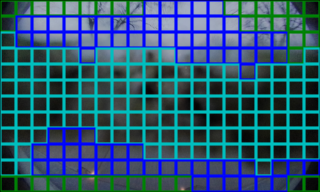}
\endminipage
\hspace{0.05mm}
\minipage{0.13\textwidth}
    \includegraphics[width=\linewidth]{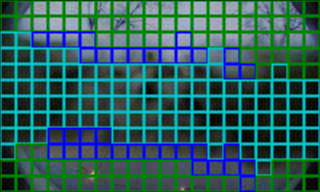}
\endminipage
\hspace{0.05mm}
\minipage{0.13\textwidth}
    \includegraphics[width=\linewidth]{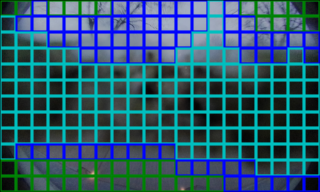}
\endminipage
\hspace{0.05mm}
\minipage{0.13\textwidth}
    \includegraphics[width=\linewidth]{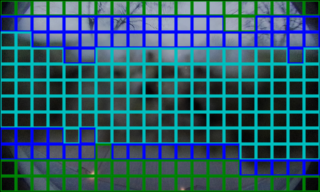}
\endminipage

\minipage{0.13\textwidth}
    \includegraphics[width=\linewidth]{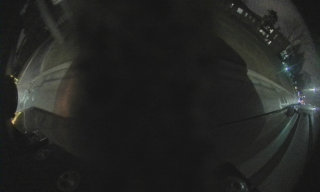}
\endminipage
\hspace{0.05mm}
\minipage{0.13\textwidth}
    \includegraphics[width=\linewidth]{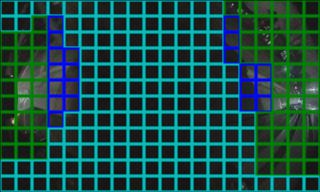}
\endminipage
\hspace{0.05mm}
\minipage{0.13\textwidth}
    \includegraphics[width=\linewidth]{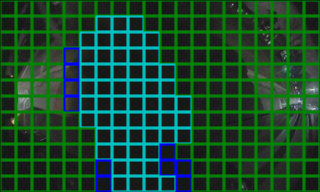}
\endminipage
\hspace{0.05mm}
\minipage{0.13\textwidth}
    \includegraphics[width=\linewidth]{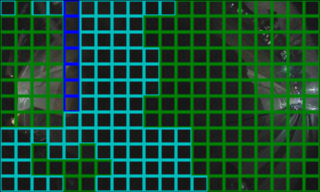}
\endminipage
\hspace{0.05mm}
\minipage{0.13\textwidth}
    \includegraphics[width=\linewidth]{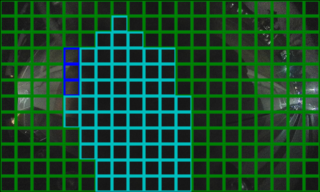}
\endminipage
\hspace{0.05mm}
\minipage{0.13\textwidth}
    \includegraphics[width=\linewidth]{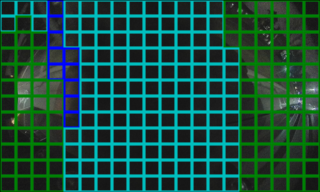}
\endminipage
\hspace{0.05mm}
\minipage{0.13\textwidth}
    \includegraphics[width=\linewidth]{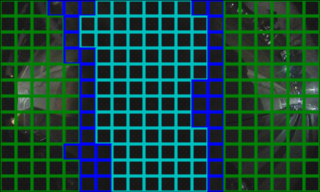}
\endminipage

\minipage{0.13\textwidth}
    \includegraphics[width=\linewidth]{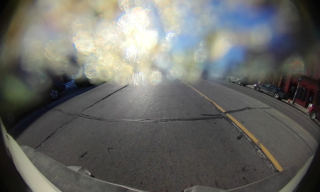}
\endminipage
\hspace{0.05mm}
\minipage{0.13\textwidth}
    \includegraphics[width=\linewidth]{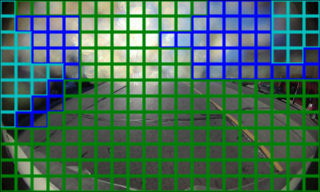}
\endminipage
\hspace{0.05mm}
\minipage{0.13\textwidth}
    \includegraphics[width=\linewidth]{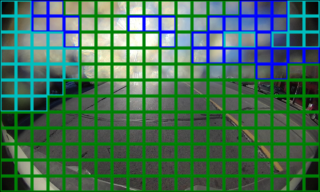}
\endminipage
\hspace{0.05mm}
\minipage{0.13\textwidth}
    \includegraphics[width=\linewidth]{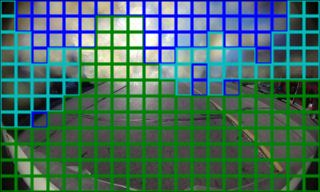}
\endminipage
\hspace{0.05mm}
\minipage{0.13\textwidth}
    \includegraphics[width=\linewidth]{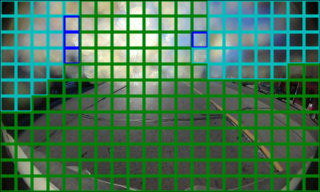}
\endminipage
\hspace{0.05mm}
\minipage{0.13\textwidth}
    \includegraphics[width=\linewidth]{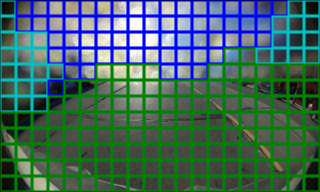}
\endminipage
\hspace{0.05mm}
\minipage{0.13\textwidth}
    \includegraphics[width=\linewidth]{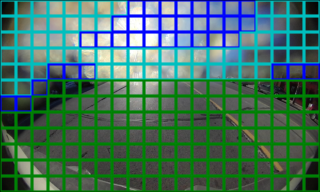}
\endminipage

\minipage{0.13\textwidth}
    \includegraphics[width=\linewidth]{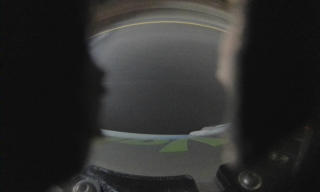}
    \hspace*{0.7cm} Input
\endminipage
\hspace{0.05mm}
\minipage{0.13\textwidth}
    \includegraphics[width=\linewidth]{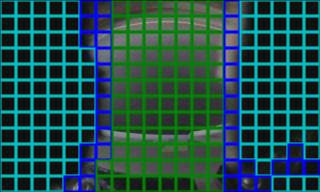}
    \hspace*{0.7cm} Net-$1$
\endminipage
\hspace{0.05mm}
\minipage{0.13\textwidth}
    \includegraphics[width=\linewidth]{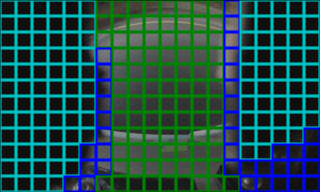}
    \hspace*{0.7cm} Net-$2$
\endminipage
\hspace{0.05mm}
\minipage{0.13\textwidth}
    \includegraphics[width=\linewidth]{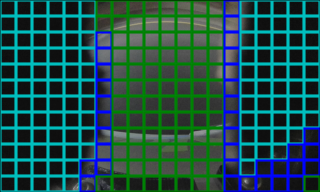}
    \hspace*{0.7cm} Net-$3$
\endminipage
\hspace{0.05mm}
\minipage{0.13\textwidth}
    \includegraphics[width=\linewidth]{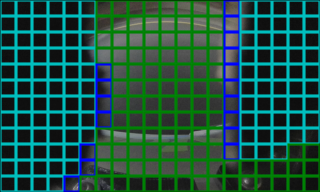}
    \hspace*{0.7cm} Net-$4$
\endminipage
\hspace{0.05mm}
\minipage{0.13\textwidth}
    \includegraphics[width=\linewidth]{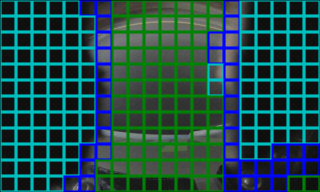}
    \hspace*{0.4cm} SoildNet
\endminipage
\hspace{0.05mm}
\minipage{0.13\textwidth}
    \includegraphics[width=\linewidth]{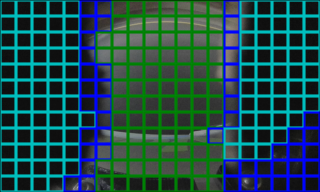}
    \hspace*{0.7cm} GT
\endminipage

\caption{Examples of $64 \times 64$ tile based soiling degradation detection output by the proposed network recommendations compared to GT (Ground Truth). From left to right: Input image, output from Net-$1$, Net-$2$, Net-$3$, Net-$4$, SoildNet, GT. \newline Color codes: Green - \textit{Clean}, Cyan - \textit{Opaque}, Blue - \textit{Transparent}. Best viewed in color.}\label{network_results}
\end{figure*}

Table \ref{network_results_classwise} summarizes the performance of all networks and the effectiveness of SoildNet is noticeable among other network propositions. The recipe of dynamic group convolution with channel reordering makes the network robust to learn better discriminative features for all classes equally and the proof is about $10$\% gain in TPR for class \textit{transparent} from the base network (Net-$1$) without degrading the performance of other classes. The results between Net-$2$ vs. Net-$3$ furnish the efficacy of channel reordering with static number of groups through out the network. The effectiveness of channel reordering with group convolution of dynamic number of groups can be seen in Net-$4$ vs. SoildNet performance. Even though Net-$2$ shows promising performance for class \textit{clean} but it fails to provide a reasonable accuracy for class \textit{transparent}. However, it is true that \textit{transparent} class has comparatively low TPR across all networks and the possible justification is that it is often confused with \textit{clean} class. Table \ref{network_results_average} further summarizes the results by computing the average of class wise accuracy for each metric. The main take away of this result is that out of $5$ standard metrics used in this experiment SoildNet outperforms other networks on $4$ metrics and thus it becomes the best proposition among all. Apart from metrics, the output of all network schemes on soiling degradation detection is demonstrated in figure \ref{network_results} where the soiling outputs are at tile level of size $64 \times 64$. In grid representation, different color codes such as green, cyan and blue are used to indicate \textit{clear}, \textit{opaque} and \textit{transparent} classes.

\begin{table}[!htb]
\centering
\renewcommand{\arraystretch}{1.5}
\scalebox{0.8}{
\begin{tabular}{||c c c c c c||}
 \hline
 - & \multicolumn{5}{c||}{Average}  \\ 
 \hline
 Network & TPR & TNR & FPR & FNR & FDR \\
 \hline\hline
 Net-$1$ & 0.7901 & 0.9406 & 0.059 & 0.2098 & 0.1891  \\ 
 Net-$2$ & 0.751 & 0.9248 & 0.0751 & 0.2488 & 0.1473 \\
 Net-$3$ & 0.8115 & 0.9497 & 0.0502 & 0.1883 & 0.1654 \\ 
 Net-$4$ & 0.7359 & 0.9269 & 0.0729 & 0.264 & \textbf{0.1278} \\
 SoildNet & \textbf{0.8277} & \textbf{0.9557} & \textbf{0.0442} & \textbf{0.1721} & 0.1907 \\ [0.25ex]
 \hline
\end{tabular}}
\vspace{0.1cm}
\caption{Comparison of average classwise accuracy between the base model (Net-$1$) and other network schemes (Net-$2$, Net-$3$, Net-$4$, SoildNet)} 
\label{network_results_average}
\end{table}



\section{Conclusion}
In this work, soiling degradation detection task, an extremely critical but relatively less explored problem has been presented in the field of autonomous driving. The solution proposed in this paper came through several interim network propositions, in particular adaptability of group convolution with static or dynamic number of groups and channel reordering in low resource environment. In this study, extensive experiment outcomes on a considerably large soiling dataset can be summarized as follows: $1)$ group convolution at all convolution layers reduces the network complexity immensely, $2)$ channel reordering can be effective to blend the features across channels and $3)$ channel reordering is more effective with group convolution with dynamic number of groups. The network schemes presented in this paper are domain agnostic and can be easily adapted in the encoder architecture for any vision tasks to be deployed on low resource platform.

{\small
\bibliographystyle{ieeetr}
\bibliography{neurips_2019}
}

\end{document}